\documentclass{article}

\usepackage[preprint]{corl_2020} % Uncomment for the camera-ready ``final'' version.

\usepackage{graphicx}
\usepackage[table]{xcolor}%
\usepackage{bbold}
\usepackage{amsmath}
\usepackage{makecell}

\title{Unsupervised Monocular Depth Learning \\in Dynamic Scenes}

\author{Hanhan Li\textsuperscript{1} \quad
Ariel Gordon\textsuperscript{1,2} \quad
Hang Zhao\textsuperscript{3} \quad
Vincent Casser\textsuperscript{3} \quad
Anelia Angelova\textsuperscript{1,2} \\

\texttt{\small\{uniqueness@google, gariel@google, hangz@waymo, casser@waymo, anelia@google\}.com} \\
\textsuperscript{1}Google Research \quad \textsuperscript{2}Robotics at Google \quad \textsuperscript{3}Waymo LLC
}

\newcommand{\etal}{\emph{et al.~}}
\newcommand{\norm}[1]{\left|#1\right|}

\begin{document}
\maketitle
%===============================================================================

\begin{abstract}
    We present a method for jointly training the estimation of depth, ego-motion, and a dense 3D translation field of objects relative to the scene, with monocular photometric consistency being the sole source of supervision. We show that this apparently heavily underdetermined problem can be regularized by imposing the following prior knowledge about 3D translation fields: they are sparse, since most of the scene is static, and they tend to be piecewise constant for rigid moving objects. We show that this regularization alone is sufficient to train monocular depth prediction models that exceed the accuracy achieved in prior work for dynamic scenes, including methods that require semantic input. \footnote{\scriptsize Code is available at \href{https://github.com/google-research/google-research/tree/master/depth_and_motion_learning}{github.com/google-research/google-research/tree/master/depth\_and\_motion\_learning}}
\end{abstract}

\keywords{Unsupervised, Monocular Depth, Object Motion} 

%===============================================================================

\section{Introduction}
Understanding 3D geometry and object motion from camera images is an important problem for robotics applications, including autonomous vehicles \cite{janai2017computer} and drones \cite{saripalli2002vision}. While robotic systems are often equipped with various sensors, depth prediction from images remains appealing in that it solely requires an optical camera, which is a very cheap and robust sensor. Object motion estimation is a nontrivial problem across various sensors.

Estimating depth and object motion in 3D given a monocular video stream is an ill-posed problem, and generally heavily relies on prior knowledge. The latter is readily provided by deep networks, that can learn the priors through training on large collections of data.

Multiple methods have been devised for providing supervision to these networks. While depth prediction networks can be supervised by sensors \cite{eigen2014depth,liu2015learning,kuznietsov2017semi}, datasets providing object 3D motion supervision for image sequences are scarce. Self-supervised methods, which rely mainly on the monocular video itself for supervision, have been attracting increasing attention recently \cite{zhou2017unsupervised, yin2018geonet,ranjan2019competitive,zou2018df,luo2018every,gordon2019depth,casser2019struct2depth,godard2018digging}, due to the virtually unlimited abundance of unlabeled video data.

\begin{figure}[h]
\vspace{0mm}
\begin{center}
 \includegraphics[width=0.98\linewidth]{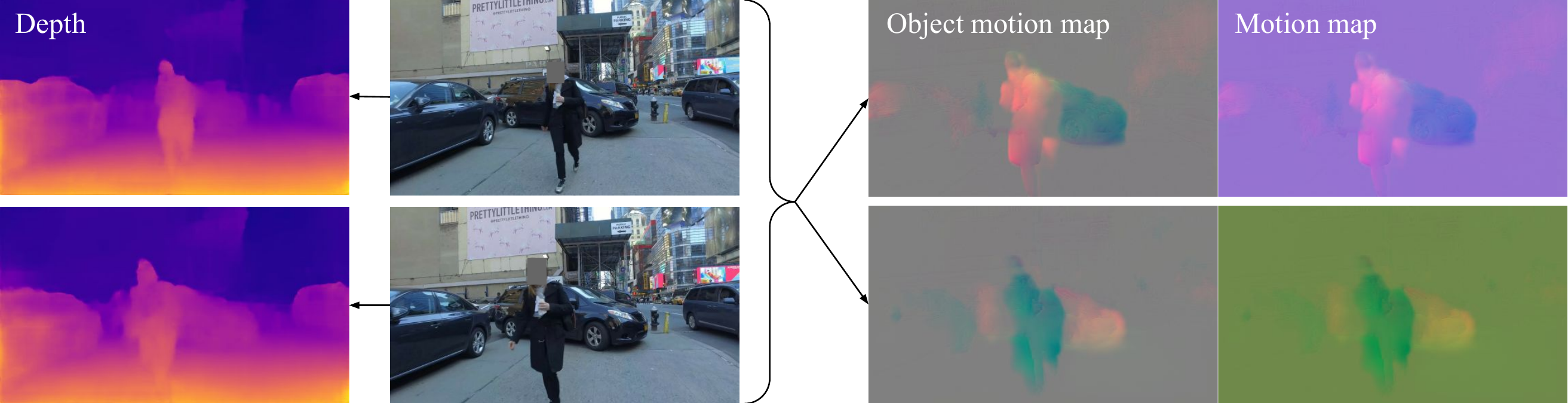}

\end{center}
  \caption{\small Depth prediction (for each frame separately) and motion map prediction (for a pair of frames), shown on a training video from YouTube. The total 3D motion map is obtained by adding the learned camera motion vector to the object motion map. Note that the motion map is mostly zero, and nearly constant throughout a moving object. This is a result of the motion regularizers used.}
\label{fig:introyt}
\end{figure}

Self-supervised learning of depth estimation is based on principles of structure from motion (SfM): When the same scene is observed from two different positions, the views will be consistent if a correct depth is assigned to each pixel, and the camera movement is correctly estimated. As such, these methods tend to suffer from many of the challenges of SfM: textureless areas, occlusions, reflections, and -- perhaps above all -- moving objects. Complete view consistency can only be achieved if the motion of every object between the capture times of the two frames is correctly accounted for.
If any point in the scene is presumed to be in motion, it carries four unknowns (depth and three motion components), which is far too many for epipolar geometry constraints to disambiguate. Self-supervised methods thus often rely on additional cues. 

One source of additional information is semantics \cite{casser2019struct2depth}. Movable objects, such as vehicles can be identified using an auxiliary segmentation model, and a network can be trained to estimate the motion of each object separately. However such techniques depend on access to an auxiliary segmentation model, capable of segmenting out all classes of movable objects to appear in the video.

Other approaches utilize different types of prior knowledge. For example, a common case of object motion in the self-driving setting is where the observing car follows another car, at approximately the same velocity. The observed car thus appears static. Godard \etal\cite{godard2018digging} propose a method that identifies this case by detecting regions that do not change between frames and excludes these regions from the photometric consistency loss. It is a very common case in the KITTI dataset, and addressing it results in significant improvements in depth estimation metrics. Yet, the method remains limited to only one specific type of object motion.

Lastly, there are approaches where optical flow is learned jointly with depth, unsupervised \cite{yang2018every}. However, stereo input is used to disambiguate the depth prediction problem.

The main contribution of this paper is a method for learning jointly depth, ego-motion and a dense object motion map in 3D from \emph{monocular video only}, where unlike prior work, our method:
\begin{itemize}
    \item Does not utilize \emph{any} auxiliary signals apart from the monocular video itself: neither semantic signals, nor stereo, nor any kind of groundtruth.
    \item Accounts for any object motion pattern that can be approximated by a rigid object translation in an arbitrary direction.
\end{itemize}

A key contribution of our paper is a novel regularization method for the residual translation fields, based on the $\frac12$ norm, which casts the residual motion field into the desired pattern described above.

In our method, a deep network predicts a dense 3D translation field (from a pair of frames). The translation field can be decomposed to the sum of background translation relative to the camera (due to ego-motion), which is constant, and an object translation field, which accounts for the motion of every point in the field of view relative to the scene. Another network predicts depth for each pixel (from each frame separately), totaling four predicted quantities per pixel. Figure~\ref{fig:introyt} illustrates the depth and translation fields. At inference time, depth is obtained from a single frame, whereas camera motion and the object translation field are obtained from a pair of frames.

Since we aim to only use a monocular video for supervision, this problem requires significant amounts of regularization. We aim to strike the right balance between regularizing sufficiently, while at the same time preserving the ability to model diverse and complex motion patterns in highly dynamic scenes. We achieve this by utilizing two observations about the nature of residual translation fields:
(1) They are \emph{sparse}, since typically most of the pixels in a frame belong to the background or static objects, and
(2) They tend to be \emph{constant throughout a rigid moving object} in 3D space.

%\footnote{Strictly speaking, this is only true for rigid objects that do not rotate. However, given sufficiently high frame rates, this assumption tends to be a good approximation.}.%object rotation is predominantly slow enough for many dynamic scenes of interest, such as streets}.
%\end{itemize}

\begin{figure}
\begin{center}
%\hspace{-0.12cm}
\includegraphics[width=0.98\linewidth]{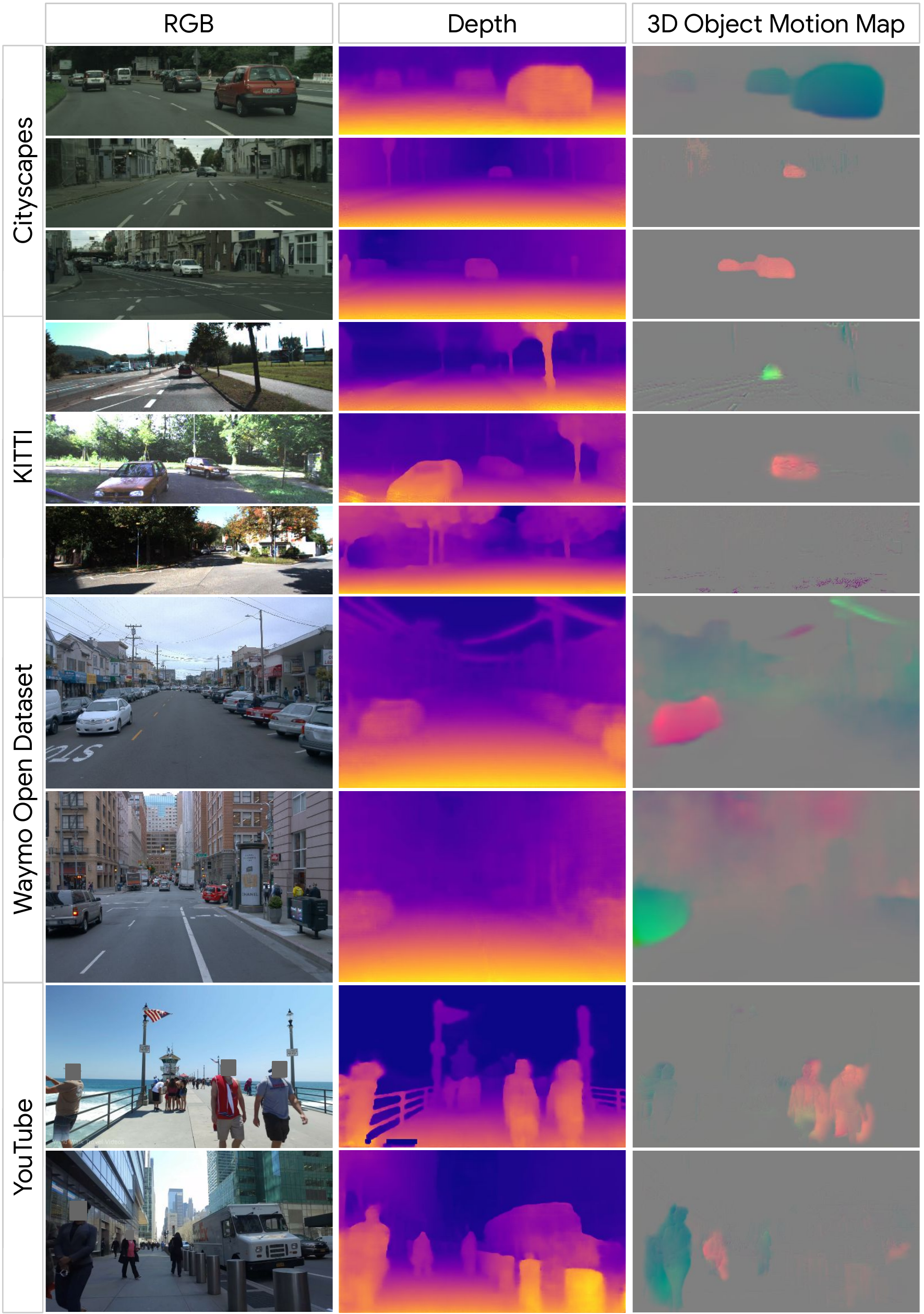}
\end{center}
   \caption{\small Qualitative results of our unsupervised monocular depth and 3D object motion map learning in dynamic scenes across all datasets: Cityscapes, KITTI, Waymo Open Dataset and YouTube.}
\label{fig:intro}
\end{figure}

We evaluate the performance of the method on four challenging datasets with dynamic scenes.
We establish new state-of-the-art results for unsupervised depth prediction on Cityscapes~\cite{Cordts2016Cityscapes} and the Waymo Open Dataset \cite{sun2019scalability}, and match the state of the art on KITTI\footnote{Since dynamic scenes are rare in the KITTI dataset, the improvements of our method on this benchmark are less pronounced.} \cite{geiger2013vision}. To further demonstrate the generality of our approach, we also train and qualitatively evaluate our model on a collection of public YouTube videos taken with hand-held cameras while walking in a variety of environments. Qualitative results from all the above datasets are shown in Figure~\ref{fig:intro}.

\section{Related Work}
\textbf{Structure from Motion and Multiview Stereo.}
Depth estimation is an important task for 3D scene understanding and robotics. Traditional computer vision approaches rely on identifying correspondences between keypoints in two or more images of the same scene and using epipolar geometry to solve for their depths~\cite{Hartley2004,snavely2006photo,schonberger2016sfm}. These methods yield sparse depth maps. They can be applied to dynamic scenes in a multi-camera setting (``multiview stereo"), or to static scenes in a single moving camera setting (``structure from motion"). Please see~\cite{ozyesil2017sfm} for a detailed survey.

\textbf{Depth Estimation.} In deep-learning based approaches \cite{eigen2014depth,liu2015learning,cao2016estimating,kuznietsov2017semi,tulsiani2017multiview}, a deep network predicts a dense depth map. These networks can be trained by direct supervision, such as through LiDAR sensors. Similar approaches are used for other dense predictions such as surface normals~\cite{eigen2015normals,Wang2015Designing}. 
More recently, deep-learning approaches have been used in conjunction with classical computer vision techniques to learn depth and ego-motion prediction~\cite{zhou2017unsupervised,garg2016unsupervised,godard2017unsupervised,ummenhofer2017demon,mahjourian2018unsupervised,yin2018geonet}. Instead of identifying keypoints in scenes and finding correspondences, deep networks predict dense depth maps and camera motion, and these are used for differentiably warping pixels from one view to another. By applying photometric consistency losses on a transformed and the corresponding reference view, a supervision signal for the model is derived.
% Already discussed and repetitive to prior work
%Garg et al.~\cite{garg2016unsupervised} introduced joint learning of depth and ego-motion. Zhou et al.~\cite{zhou2017unsupervised} used a fully- differentiable loss that allowed learning depth and ego-motion, end to end, form unlabeled video frame sequences. 
A lot of progress in monocular depth learning has followed~\cite{ummenhofer2017demon,Yang2017unsupervised,mahjourian2018unsupervised,yin2018geonet,wang2018learning,casser2019struct2depth,li2019movingpeople}, and some work proposed to use stereo inputs for training, with the purpose of producing more accurate monocular depth at inference~\cite{godard2017unsupervised,ummenhofer2017demon,yang2018every,wang2018learning,zhan2018unsupervised}. Alternative approaches learn to produce the stereo disparity~\cite{kendall2017end,khamis2018stereonet,yao2018mvsnet}, or apply depth completion from an online depth sensor~\cite{mal2018sparse,zhang2018deep}.

\textbf{Depth and Motion.} When the camera and objects move with respect to the scene, enforcing consistency across views requires estimating motion of both the camera as well as individual objects. 
As estimating depth in dynamic scenes is very challenging, many approaches have used multiple views to obtain depth~\cite{ranftl2016dense,kumar2017monocular}.
Several approaches have recently been proposed for simultaneously learning depth, camera motion, and object motion from monocular videos. Yin \etal \cite{yin2018geonet} and Zhou \etal \cite{zou2018df} learn to jointly predict depth, ego-motion and optical flow. The first stage of their method estimates depth and camera motion, and thus the optical flow induced by camera motion. A second stage estimates the residual optical flow due to object motion relative to the scene. The residual flow is used to mask out moving objects and to reason about occlusions. Luo \etal \cite{luo2018every} also jointly optimize depth, camera motion, and optical flow estimation, using a holistic motion parser. Stereo input is used to disambiguate the depth prediction training. Casser \etal \cite{casser2019struct2depth,casser2019unsupervised} estimate the motion of objects in the scenes, with the assistance of pre-trained segmentation models, leading to significant improvement in depth estimation for moving objects. Gordon \etal \cite{gordon2019depth} introduce a differentiable way of handling occlusions but still require an auxiliary segmentation model for moving objects. Li \etal \cite{li2019movingpeople} are able to handle scenes with moving people but require human-segmentation masks and relies on a dataset with ``static" people, observed from multiple views. Godard \etal \cite{godard2018digging} address object motion by detecting cases where an object is static with respect to the camera (i.~e.~moving at the same speed), which addresses a very common case of object motion observed from vehicles in traffic.
Our work provides a more general way of performing unsupervised depth learning in dynamic scenes; unlike previous work, we do not need to segment out objects in order to estimate their motion, nor do we assume stereo data. Unlike prior work that uses residual optical flow to reason about moving objects and occlusions, our method directly  regularizes motion in 3D, which turns out to lead to better depth prediction accuracy. % Lastly our method for object motion estimation is not limited to objects moving at the same speed as the camera.

%------------------------------------------------------------------------
\section{Method}
In our approach, depth, ego-motion, and object motion are learned simultaneously from monocular videos using self-supervision. 
We use pairs of adjacent video frames ($\mathbf{I_a}$ and $\mathbf{I_b}$) as training data. Our depth network predicts a depth map $D(u, v)$ (where $u$ and $v$ are the image coordinates) at the original resolution from a single image, and we apply it independently on each of the two frames. 
The two depth maps are concatenated with $\mathbf{I_a}$ and $\mathbf{I_b}$ in the channel dimension and are fed into a motion prediction network (Figure~\ref{fig:introyt}). The latter predicts a 3D translation map $T_{\rm obj}(u, v)$ at the original resolution for the moving objects and a 6D ego-motion vector $M_{\rm ego}$. $M_{\rm ego}$ consists of the 3 Euler angles that parameterize the 3D rotation matrix  $\mathbf{R}$ and elements of the 3D translation vector $T_{\rm ego}$. The object motion relative to the camera is defined as rotation $\mathbf{R}$ followed by a translation
\begin{equation}
T(u, v)=T_{\rm obj}(u, v)+T_{\rm ego}.
\label{translation_ansatz}
\end{equation}

\begin{figure}[h]
\begin{center}
     \includegraphics[width=1.0\linewidth]{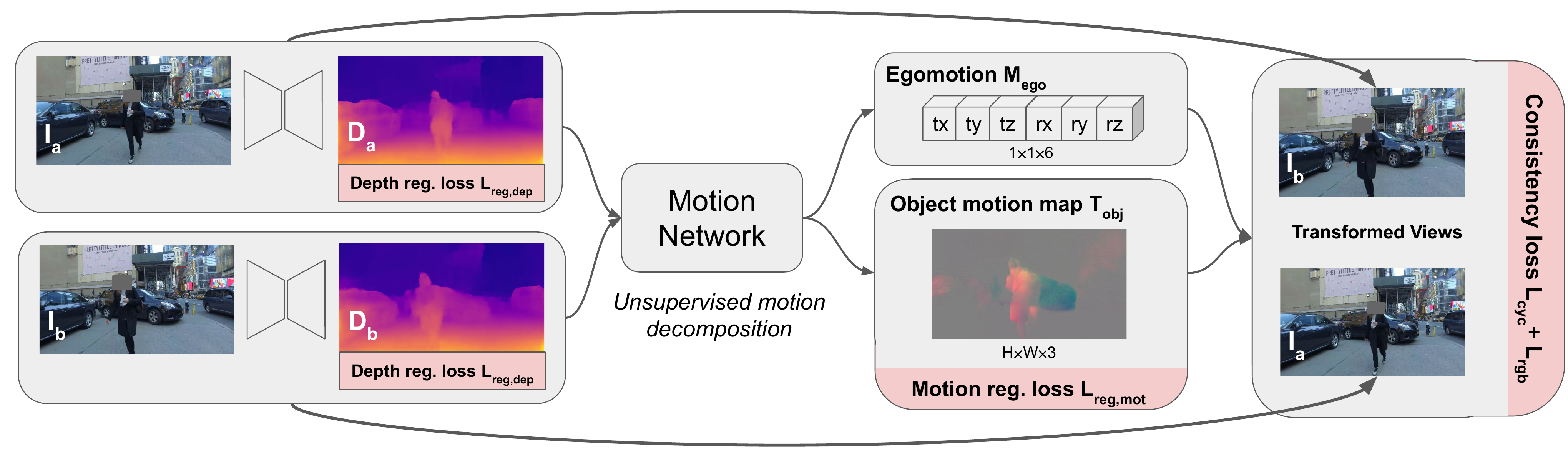}
\end{center}
\caption{\small Overall training setup. A depth network is independently applied on two adjacent RGB frames, $\mathbf{I_a}$ and $\mathbf{I_b}$, to produce the depth maps, $\mathbf{D_a}$ and $\mathbf{D_b}$. The depth maps together with the two original images are fed into the motion network, which decomposes the motion into a global ego-motion estimate $M_{ego}$ and a spatial object motion map $T_{obj}$. Given motion and depth estimates, a differentiable view transformer allows transitioning between them. Losses are highlighted in red. For example, we use a motion regularization loss (Section~\ref{subsubsec:motion_reg}) on the motion map, and a motion cycle consistency loss and a photometric consistency loss (Section~\ref{subsubsec:ConsistencyRegularization}). %This is performed twice, reverting the places of the first and second images, i.e., the input image $\mathbf{I_a}$ is switched with $\mathbf{I_b}$, and the input depth $D_a$ is switched with $D_b$. 
%A composite of regularization losses is imposed on the network predictions. 
At inference time, a depth map is obtained from a single frame, whereas a 3D motion map and ego-motion are obtained from two consecutive frames.}
\label{fig:architecture}
\end{figure}

We propose new motion regularization losses on $T(u, v)$ (Sec.~\ref{subsubsec:motion_reg}) which facilitate training in highly dynamic scenes. The overall training setup is shown in Figure~\ref{fig:architecture}.
Figure \ref{fig:intro} visualizes examples of the learned motion $T_{\rm obj}(u, v)$ and disparity $d(u, v) = 1/D(u, v)$ per frame.

\subsection{Depth and Motion Networks}
Our depth network is an encoder-decoder architecture, identical to the one in Ref.~\cite{zhou2017unsupervised}, with the only difference that we use a softplus activation function for the depth, $z(\ell)=\log(1+\mathrm e^\ell)$. In addition, randomized layer normalization \cite{gordon2019depth} is applied before each $relu$ activation in the network.

The input to the motion network is a pair of consecutive frames, concatenated along the channel dimension. The motion prediction is similar to the one in Ref.~\cite{gordon2019depth}, with the difference that each input image has four channels: The three RGB channels, and the predicted depth as the fourth channel. The rationale is that having depth as a fourth channel, even if predicted rather than exactly measured, provides helpful signals for the task of estimating motion in 3D.

\subsection{Losses}
Training is driven by a number of self-supervised losses as is standard in monocular or stereo-based unsupervised depth and ego-motion learning. The total loss is the sum of three components, the motion regularization, the depth regularization, and the consistency regularization. The motion regularization and consistency regularization are applied twice on the frame pair, and the frame order is reversed in the second application. The depth regularization is applied independently on each one of pair. We describe the losses below, starting with the motion regularization losses, which are a key contribution of this work. 

\subsubsection{Motion Regularization}
\label{subsubsec:motion_reg}

The regularization $L_{\rm reg,mot}$ on the motion map $T_{\rm obj}(u, v)$ consists of the \textbf{\textit{group  smoothness loss}} $L_{g1}$ and the $L_{1/2}$ \textbf{\textit{ sparsity loss}}. The group smoothness loss $L_{g1}$ on $T_{\rm obj}(u, v)$ minimizes changes within the moving areas, encouraging the motion map to be nearly constant throughout a moving object. This is done in anticipation that moving objects are mostly rigid.
It is defined as: 
\begin{equation}
    L_{g1}[T(u, v)]= \sum_{i\in\{x,y,z\}}\iint \sqrt{\big(\partial_u T_{i}(u, v)\big)^2 + \big(\partial_v T_{i}(u, v)\big)^2} \mathrm{d}u\mathrm{d}v
\end{equation}
The $L_{1/2}$ sparsity loss on $T_{\rm obj}(u, v)$ is defined as:
\begin{equation}
    L_{1/2}[T(u, v)]= 2\sum_{i\in\{x,y,z\}}\langle\left| T_i \right|\rangle\iint \sqrt{1+ \left| T_{i }(u, v) \right| /\langle\left| T_i \right|\rangle } \mathrm{d}u\mathrm{d}v
\label{eqa:l1_2}
\end{equation}
where $\langle\left| T_i \right|\rangle$ is the spatial average of $\left|T_i(u, v)\right|$. The coefficients are designed in this way so that the regularization is self-normalizing. In addition, it approaches $L_1$ for small $T(u, v)$, and its strength becomes weaker for larger $T(u, v)$. We visualize its behavior in the Supplemental Material. 
Overall, the $L_{1/2}$ loss encourages more sparsity than the $L_1$ loss.

The final motion regularization loss is a combination of the above losses:
\begin{equation}
L_{\rm reg,mot} = \alpha_{\rm mot} L_{g1}[T_{\rm obj}(u, v)] + \beta_{\rm mot} L_{1/2}[T_{\rm obj}(u, v)] 
% L_2(\theta) +
\end{equation}
where $\alpha_{\rm mot}$ and $\beta_{\rm mot}$ are hyperparameters.

Strictly speaking, a piecewise-constant $T_{\rm obj}(u, v)$ can describe any scene where objects are moving in pure translation relative to the background. However, when objects are rotating, the residual translation field is generally not constant throughout them. Since fast rotation of objects relative to the background is uncommon, especially in road traffic, we expect the piecewise-constant approximation to be appropriate.

\subsubsection{Depth Regularization}
We apply a standard edge-aware smoothness regularization on the disparity maps $d(u, v)$ as described in Godard \etal \cite{godard2017unsupervised}. In other words, the regularization is weaker around pixels where color variation is higher:
\begin{equation}
L_{\rm reg,dep} = \alpha_{\rm dep} \iint (|\partial_u d(u, v)| \mathrm{e}^{-\norm{\partial_u \mathbf{I}(u, v)}}+|\partial_v d(u, v)| \mathrm{e}^{-\norm{\partial_v \mathbf{I}(u, v)}} ) \mathrm{d}u\mathrm{d}v
\end{equation}
where $\alpha_{\rm dep}$ is a hyperparameter.

\subsubsection{Consistency Regularization}
\label{subsubsec:ConsistencyRegularization}
The consistency regularization is the sum of the motion cycle consistency loss $L_{\rm cyc}$ and the occlusion-aware photometric consistency loss $L_{\rm rgb}$. $L_{\rm cyc}$ encourages the forward and backward motion between any pair of frames to be the opposite of each other. 
\begin{equation}
L_{\rm cyc} = \alpha_{\rm cyc} \frac{ \lVert\mathbf{R}\mathbf{R}_{\rm inv}-\mathbb{1}\rVert^2} { \lVert\mathbf{R}-\mathbb{1}\rVert^2 + \lVert\mathbf{R}_{\rm inv}-\mathbb{1}\rVert^2}
+ \beta_{\rm cyc} \iint \frac{ \lVert\mathbf{R}_{\rm inv} T(u, v) + T_{\rm inv}(u_\mathrm{warp}, v_\mathrm{warp})\rVert^2} {\lVert T(u, v) \rVert^2+ \lVert T_{\rm inv}(u_\mathrm{warp}, v_\mathrm{warp})\rVert^2 } \mathrm{d}u\mathrm{d}v
\end{equation}
where the `inv' subscript indicates that the same quantity was obtained with the input frames reversed in order. $\alpha_{\rm cyc}$ and $\beta_{\rm cyc}$ are hyperparameters.

$L_{\rm rgb}$ encourages photometric consistency of corresponding areas in the two input frames. Similar to prior works, it is a sum of a L1 loss and a SSIM structural similarity loss in the RGB space.
\begin{equation}
L_\mathrm{rgb} = \alpha_{\rm rgb} \iint \norm{\mathbf{I}(u,v)-\mathbf{I}_\mathrm{warp}(u,v)}\mathbb{1}_{D(u,v)>D_\mathrm{warp}(u,v)} \mathrm{d}u\mathrm{d}v
+ \beta_{\rm rgb} \frac{1-\mathrm{SSIM}(\mathbf{I}, \mathbf{I}_\mathrm{warp} )}{2}
\end{equation}
where $\mathbf{I}$ and $\mathbf{I}_\mathrm{warp}$ are the original image and warped image, $\mathbb{1}_{D(u,v)>D_\mathrm{warp}(u,v)}$ is a mask introduced in Gordon \etal ~\cite{gordon2019depth} to address occlusions. $\alpha_{\rm rgb}$ and $\beta_{\rm rgb}$ are hyperparameters.

The frame warping is calculated using the camera intrinsic matrix $\mathbf{K}$, the rotation matrix $\mathbf{R}$, and the camera translation vector $T$ as $z'p' = \mathbf{K}\mathbf{R}\mathbf{K}^{-1}zp +  \mathbf{K} T$. Here $p$ and $z$ are the homogeneous pixel coordinates and the depth, and their primed counterparts are warped ones in the new frame.

\section{Experiments}
In this section, we present results on a variety of datasets, including Cityscapes, KITTI, Waymo Open Dataset and a collection of videos taken with moving cameras from YouTube. For all our experiments, the encoder part of the depth network is initialized from a network pretrained on ImageNet \cite{deng2009imagenet}. The camera intrinsic matrices are provided in all datasets except for Youtube videos, where they are learned as part of our models. The hyperparameter values we used are the same for all experiments and are given in the supplementary material.

The inference time of our depth prediction model is about 5.3ms per frame of resolution 480x192, at a batch size of 1 on a NVIDIA V100 GPU (unoptimized), which is equivalent to roughly 190 frames per second. 
As far as we are aware, this is among the fastest methods.
\subsection{Cityscapes}

The Cityscapes \cite{Cordts2016Cityscapes} dataset is an urban driving dataset, which is quite challenging for unsupervised monocular depth estimation, because of the prevalence of dynamic scenes. As a result, not many works published results on this dataset, with few exceptions~\cite{casser2019struct2depth,gordon2019depth,pilzer2018unsupervised}. We use standard evaluation protocols as in prior work~\cite{casser2019struct2depth,pilzer2018unsupervised}. For training we combine the densely and coarsely annotated splits to obtain 22,973 image-pairs. For evaluation, we use the 1,525 test images. 
The evaluation uses the code and methodology from Struct2Depth~\cite{casser2019struct2depth}.

\begin{table*}[h]
\begin{center}
\resizebox{\columnwidth}{!}{%

  \begin{tabular}{|l|c||c|c|c|c||c|c|c|}
  \hline
  Method   &\small{Uses semantics?} & \cellcolor{red!15} Abs Rel& \cellcolor{red!15}Sq Rel & \cellcolor{red!15}RMSE & \cellcolor{red!15}RMSE log & \cellcolor{green!15}$\delta < 1.25$ & \cellcolor{green!15}$\delta < 1.25^2$ & \cellcolor{green!15}$\delta < 1.25^3$ \\
  \hline 
  \hline
  Struct2Depth \cite{casser2019struct2depth}  &Yes & 0.145 &1.737 &7.28 & 0.205 &0.813 & 0.942 & 0.978 \\ 
  Gordon \cite{gordon2019depth}  &Yes &0.127 & 1.33 & \textbf{6.96} & 0.195 &0.830  & \textbf{0.947} & \textbf{0.981} \\ 

  \hline
  \hline

  Pilzer \cite{pilzer2018unsupervised} &\textbf{No} &0.440& 6.04 & 5.44 & 0.398 & 0.730 & 0.887 & 0.944\\
  \hline

  Ours &\textbf{No} & \textbf{0.119} &\textbf{1.29} & \textbf{6.98}  &\textbf{0.190} &\textbf{0.846} & \textbf{0.952}&\textbf{ 0.982}\\
  \hline
  \end{tabular}%
  }
\end{center}
\caption{\small Performance comparison of unsupervised single-view depth learning approaches, for models trained and evaluated on Cityscapes using the standard split. The depth cutoff is 80m. Our model uses a resolution of $416 \times 128$  for input/output. The `uses semantics' column indicates whether the corresponding method requires a pretrained mask network to help identify moving objects. Our approach does not use semantics information. `Abs Rel', `Sq Rel', `RMSE', and `RMSE log' denotes mean absolute error, squared error, root mean squared error, and root mean squared logarithmic error respectively. $\delta < x$ denotes the fraction with the ratio between groundtruth and prediction values between $x$ and $1/x$. For the red metrics, lower is better; for the green metrics, higher is better.}
\label{tab:cityscapes}
\end{table*}

\begin{table}[h]
{\footnotesize

\begin{center}
  \begin{tabular}{|l|c|c|c|c|c|c|}
  \hline
  Method   & \cellcolor{red!15} Abs Rel& \cellcolor{red!15}Sq Rel & \cellcolor{red!15}RMSE & \cellcolor{red!15}RMSE log \\
  \hline 
  \hline 
  {Ours, $L_{1/2}$, without depth prediction inputs} & 0.125 &1.41 & 7.39  &0.200\\
  \hline
  {Ours, $L_1$ instead of $L_{1/2}$} & 0.125 &1.37 & 7.33  &0.199\\
  \hline 
  {Ours, $L_{1/2}$} & \textbf{0.119} &\textbf{1.29} & 6.98  &0.190\\
  \hline
   {Ours, $L_{1/2}$, with mask} & \textbf{0.119} &1.36 & \textbf{6.89}  &\textbf{0.188}\\
  \hline
  \end{tabular}
\end{center}
}
\caption{\small Ablation study on Cityscapes. 
 $L_{1/2}$ is the loss term defined in Eq. \ref{eqa:l1_2}. `with mask' means using a pretrained segmentation model to identify regions of potentially moving objects, as in prior work~\cite{casser2019struct2depth}.}
\label{tab:cityscapes_ablation}
\end{table}

Table \ref{tab:cityscapes} compares the performance of our approach with prior works which reported results on Cityscapes.
Our method is able to outperform all prior methods except for the metric (RMSE) compared to Ref.~\cite{gordon2019depth}. However the latter method uses semantic cues.

Table \ref{tab:cityscapes_ablation} shows an ablation study on this dataset, as it contains many dynamic scenes. If we do not feed the depth predictions into the motion network, performance deteriorates - `Abs Rel' increases by about $0.006$. If we use the $L_1$ loss instead of the $L_{1/2}$ loss to regularize the object motion map, `Abs Rel' also increases by about $0.006$. We also tried augmenting our approach with a detection model pretrained in the COCO \cite{lin2014microsoft} dataset. This model produces a mask for each image which is applied onto the object motion map. However, the detection model does not cause noticeable improvements for depth estimation.

%%%%%%%%%%%%%%%%%%%%%%%%%%%%%%%%%%%%%%%%%%%%
\subsection{KITTI}

The KITTI  \cite{geiger2013vision} dataset is collected in urban environments and is a popular benchmark for depth and ego-motion estimation. It is accompanied with LiDAR data, which is used for evaluation only. While KITTI only has a small number of dynamic scenes, it is a very common dataset for evaluating depth models. We follow the Eigen split with 22,600 training image pairs and 697 evaluation pairs. We use the standard evaluation protocol, established by Zhou \etal~\cite{zhou2017unsupervised} and adopted by many subsequent works. Table \ref{tab:kitti} shows that the performance of our model is on par with the state of the art. 

\begin{table*}[h]
\begin{center}
\resizebox{\columnwidth}{!}{%
  \begin{tabular}{|l|c||c|c|c|c||c|c|c|}
  \hline
  Method  & \small{Uses semantics?} & \cellcolor{red!15} Abs Rel& \cellcolor{red!15}Sq Rel & \cellcolor{red!15}RMSE & \cellcolor{red!15}RMSE log & \cellcolor{green!15}$\delta < 1.25$ & \cellcolor{green!15}$\delta < 1.25^2$ & \cellcolor{green!15}$\delta < 1.25^3$ \\
  \hline 
  \hline
  Struct2Depth \cite{casser2019struct2depth}   &Yes & 0.141  & 1.026  & 5.291 &0.2153 &0.8160 &0.9452 &0.9791 \\
  Gordon \cite{gordon2019depth}  &Yes & \textbf{0.128} & \textbf{0.959} & 5.23 & 0.212 &  0.845 & 0.947 & 0.976\\
  \hline
  \hline
  % Mahjourian \cite{mahjourian2018unsupervised}  & \textbf{No} & 0.163 & 1.240 & 6.220 & 0.250 &0.762 &0.916 &0.968\\ 
%   LEGO \cite{yang2018lego}  & \textbf{No} &0.162 &1.352 &6.276 & 0.252 &0.783& 0.921 &0.969 \\
  % GeoNet \cite{yin2018geonet}  & \textbf{No} &0.155 &1.296 &5.857& 0.233 &0.793 &0.931 &0.973  \\
  % DDVO \cite{wang2018learning} & \textbf{No} &0.151 &1.257 &5.583 & 0.228 &0.810 &0.936 &0.974 \\
  Yang \cite{luo2018every}  &\textbf{No} &0.141 &1.029 &5.350 &0.216 &0.816 &0.941 &0.976\\
  Bian \cite{bian2019unsupervised}   &\textbf{No} &0.137 &1.089 &5.439 &0.217 &0.830 &0.942 &0.975\\
  Godard \cite{godard2018digging}  & \textbf{No}&\textbf{0.128} &1.087 &5.171 & \textbf{0.204} & \textbf{0.855} & \textbf{0.953} &\textbf{0.978} \\
  \hline

  Ours  &\textbf{No} & 0.130 & \textbf{0.950} & \textbf{5.138} &0.209 & 0.843 & 0.948& \textbf{0.978}\\
  \hline
  \end{tabular}%
}
\end{center}
\caption{\small Performance comparison of unsupervised single-view depth learning approaches, for models trained and evaluated on KITTI using the Eigen Split. The depth cutoff is 80m. All results in the table (including ours) are for a resolution of $416 \times 128$ for input/output. Please refer to Table \ref{tab:cityscapes} for more details on the table schema.}
\label{tab:kitti}
\end{table*}

%%%%%%%%%%%%%%%%%%%%%%%%%%%%%%%%%%%%%%%%%%%%
\subsection{Waymo Open Dataset}
The Waymo Open Dataset~\cite{sun2019scalability} is currently one of the largest and most diverse publicly released autonomous driving datasets. Its scenes are not only dynamic but also comprise nighttime driving and diverse weather conditions. We experiment on this dataset to showcase the generality of our method. For training, we take $100,000$ image-pairs from $1,000$ front camera video sequences. For evaluation, we use $1,500$ image-pairs from $150$ sequences. Ground truth depth from LiDAR is only used for evaluation. Since previous unsupervised depth approaches did not publish results on this dataset, we compare using available open-source code of state-of-the-art methods, capable of handling dynamic scenes.

Table \ref{tab:waymo} shows the performance of our model in comparison with prior works, both of which require semantics cues. Our model, which does not need any masks, outperforms previous works that additionally use object masks.

We also apply object masks to our method (last row), and observe further improvements comparing to the maskless setting (second last row), showing the effectiveness of our proposed regularization with or without mask. Visual examples are shown in Figure~\ref{fig:intro} and in the Supplemental Material. As shown, our approach is successfully handling these challenging scenes.

\begin{table}[h]
{\footnotesize

\begin{center}
  \begin{tabular}{|l|c|c|c|c|}
  \hline
  Method   & \cellcolor{red!15} Abs Rel& \cellcolor{red!15}Sq Rel & \cellcolor{red!15}RMSE & \cellcolor{red!15}RMSE log \\
  \hline 
  \hline
  {Open-source code from \cite{casser2019struct2depth}, with Mask} & 0.180 & 1.782 & 8.583  & 0.244 \\
  \hline
  {Open-source code from \cite{gordon2019depth}, with Mask} & 0.168 & 1.738 & 7.947  & 0.230 \\
  \hline
  {Ours, without Mask} & \textbf{0.162} & \textbf{1.711} & \textbf{7.833}  & \textbf{0.223} \\
  \hline
   {Ours, with Mask} & \textbf{0.157} & \textbf{1.531} & \textbf{7.090}  & \textbf{0.205} \\
  \hline
  \end{tabular}
\end{center}
}
\caption{\small Performance on the Waymo Open Dataset. Our approach outperforms prior work, even though it does not require masks (top portion). With masks, it performs even better (bottom).}
\label{tab:waymo}
\end{table}

The supplementary material includes videos of our method running on validation sequences of the Waymo Open Dataset and Cityscapes.

%%%%%%%%%%%%%%%%%%%%%%%%%%%%%%%%%%%%%%%%%%%%
\subsection{YouTube dataset}
To demonstrate that depth can be learned from videos in the wild, we randomly picked a collection of videos on YouTube taken with handheld cameras while walking. The videos contain many different scenes across quite a few cities, and they feature objects in a wide range of depths. Because the videos are taken with unknown cameras, we simultaneously learn an intrinsic matrix per video. An intrinsic matrix is parameterized as a set of four trainable variables: two for the focal lengths and two for the optical centers \cite{gordon2019depth}. Visual depth maps are shown in Figure~\ref{fig:intro} and in the supplemental material.

\section{Conclusions}

This paper presents a novel unsupervised method for depth learning in highly dynamic scenes, which jointly solves for 3D motion maps and depth maps. Our model can be trained on unlabeled monocular videos without requiring any auxiliary semantic information. Our method is conceptually very simple, as we use end-to-end differentiable losses that encourage photometric consistency, motion smoothness, and motion sparsity. The main limitation is that object rotation and deformation is not explicitly handled, and that camera movement needs to be present to receive learning signals. We demonstrate the efficacy of the proposed approach on Cityscapes, KITTI, the Waymo Open Dataset, and YouTube data. For datasets rich in dynamic scenes, we outperform prior depth estimation benchmarks, including ones that utilize semantic cues.

% \bibliography{CoRL_review}

\end{document}